# Imputation of missing sub-hourly precipitation data in a large sensor network: a machine learning approach

Benedict Delahaye Chivers[1], John Wallbank[2], Steven J. Cole[2], Ondrej Sebek[2], Simon Stanley[2], Matthew Fry[2], Georgios Leontidis[1,3]*

[1]*School of Computer Science, University of Lincoln, LN6 7TS, Lincoln, UK*

[2] *Water Resources, UK Centre for Ecology and Hydrology, OX10 8BB, Wallingford, UK*

[3]*Department of Computing Science, University of Aberdeen, AB24 3UE, Aberdeen, UK*

**Corresponding author. Email address: georgios.leontidis@abdn.ac.uk*

**Abstract**

Precipitation data collected at sub-hourly resolution represents specific challenges for missing data recovery by being largely stochastic in nature and highly unbalanced in the duration of rain vs non-rain. Here we present a two-step analysis utilising current machine learning techniques for imputing precipitation data sampled at 30-minute intervals by devolving the task into (a) the classification of rain or non-rain samples, and (b) regressing the absolute values of predicted rain samples. Investigating 37 weather stations in the UK, this machine learning process produces more accurate predictions for recovering precipitation data than an established surface fitting technique utilising neighbouring rain gauges. Increasing available features for the training of machine learning algorithms increases performance with the integration of weather data at the target site with externally sourced rain gauges providing the highest performance. This method informs machine learning models by utilising information in concurrently collected environmental data to make accurate predictions of missing rain data. Capturing complex non-linear relationships from weakly correlated variables is critical for data recovery at sub-hourly resolutions. Such pipelines for data recovery can be developed and deployed for highly automated and near instantaneous imputation of missing values in ongoing datasets at high temporal resolutions.

*Keywords:* machine learning, data imputation, gradient boosted trees, environmental sensor networks, precipitation, soil moisture

## 1. Introduction

Precipitation data is of critical importance across multiple lines of enquiry, informing statistical models and analysis relating to weather forecasting, extreme weather events, climate change, water-resource management, droughts, flooding, agricultural impact, and hydroelectric power. Historical rainfall data can reveal long term trends in environmental hydrological issues with real-time data input allowing for immediate forecasting of future conditions. Distributed networks of rain gauges are typically used to provide precipitation data at the earth's surface at varying temporal resolutions and can cover large geographical areas (Kidd, 2001). As is the case in many databases, particularly those utilising physical sensors, the problem of missing data arises. Missing data can be a result of sensor failure, data storage/transmission failure, or post-collection quality control procedures resulting in removal of identified problem data (Blenkinsop et al., 2017). Missing data in precipitation databases represents a serious limitation for the effective use of the data. Given the global scale and importance of precipitation and meteorological data (Sun et al., 2018), developing solutions to missing data is of paramount importance for maximising information gain.

Precipitation data presents specific problems for recovering missing values. In the temperate oceanic climate of Western Europe, rainfall is more evenly dispersed throughout the year compared to other regions (Kottek et al., 2006). Rain has both seasonal and diurnal components (Darwish et al., 2018), however, these are not consistent across geographical regions (e.g. a north-south divide of the UK mainland, Blenkinsop et al., 2017). Similarly, distribution of rain is not equal across regions and is dependent on the class of rain exhibited. Rain events are commonly classified as being 'convective' or 'stratiform' in nature (Houze Jr, 1997), with convective rain events typically being of short duration and occurring over a small geographical area. In contrast, stratiform rain can present over an area hundreds of kilometres wide (Blackburn et al., 2008). In the UK, observation of real-time precipitation readings and historical databases of rain gauge readings reveals that it rains ~10% of the time. Rain regularly occurs in isolated single time-samples when sampled at sub-hourly resolutions indicating many short bouts of rain. Thus, time-series precipitation data is lower bounded at zero and highly zero-inflated which can reduce performance of standard regression techniques when trying to recover missing data (as in Ancelet et al., 2010). Given the trajectory towards higher resolution in both spatial and temporal factors in climatological modelling (Arandia et al., 2014;), the requirements for high quality data at hourly, or higher, sampling frequencies are becoming increasingly common.

A common solution to missing data in analysis across all fields is case-removal, whereby samples or periods of records which contain missing data in the dataset are removed prior to analysis to ensure complete data (as in Blenkinsop et al., 2015). However, omission of examples with missing data both introduces bias (Myers, 2011) and the loss of information adversely affects predictive or explanatory power of models in most cases (White and Carlin, 2010).

Recovering missing rain gauge data can be achieved as simply as using analogous data from neighbouring gauges as a replacement, although correlation between gauges is not consistent at small distances (Kamaruzaman et al., 2017), and will decrease with increasing distance (Blenkinsop et al., 2017). Similarly, multiple neighbouring sensors can be used with distance-based weighted interpolation (Cole and Moore, 2008; Teegavarapu et al., 2009). However, complex and nonlinear spatio-temporal relationships in rain events across gauges may be difficult to identify from neighbouring gauge data due to potentially large distances between them (Tang et al., 2013), especially at high temporal resolution (hourly sampling or less). Many multivariate modelling techniques have been proposed for missing data imputation, including expectation maximisation, hot-deck imputation, least squares regression, predictive mean matching, nearest neighbour techniques, support vector regression, decision tree techniques, gradient boosting, and artificial neural networks (Dempster et al., 1977; Troyanskaya et al., 2001; Bø et al., 2004; Wang et al., 2006; Andridge and Little, 2010; Honaker et al., 2011; Templ et al., 2011; Bertsimas et al., 2017; Körner et al., 2018). As universal function approximators, supervised machine learning (ML) techniques are suitable for capturing the nonlinear relationships amongst distributed rain gauges and, importantly, related sensor data.

Related data can be concurrently recorded with precipitation in certain sensor installations, or recorded in external sensor networks, and includes meteorological variables which will correlate to some degree with precipitation (Shahin et al., 2014). Indeed, at sub-hourly sampling, the correlation of precipitation to meteorological variables is expected to be lower. Spatial coherence of air pressure is much more coherent than that of precipitation over large distances (100 – 500 km) between measurement sites (Thompson, 1999) and will not drastically change on a sub-hourly basis in agreement with single-point recordings of precipitation across gauges, particularly in short term convective events. Similarly, soil moisture measurements will be affected by precipitation, but will not exhibit clear peaks in association with observed rain events (Nash et al., 1991). Considering the stochastic nature of rain events, and the low correlation with other variables across short time scales, it is unknown how well predictive analyses will perform in recovering missing precipitation data at sub-hourly temporal resolution.

While many studies have addressed the prediction of precipitation data for forecasting (Devi et al., 2016), and use in context-specific hydrological modelling (Lee and Kang, 2015), fewer have approached the problem of recovering missing data for database completion, especially at sub-daily resolution. In contrast to forecasting, a benefit of missing data recovery is that data both before and after the sample of missing data can be used to leverage information. Many studies use spatial interpolation methods from neighbouring gauges (e.g. Teegavarapu, 2009), despite numerous limitations which can result in over- or under-estimation of precipitation, depending on observations (discussed in Teegavarapu, 2009).

One method to deal with zero-inflated data is to first model the occurrence of zeros as a binary condition of rain or no rain, allowing subsequent analysis of magnitude to be applied to only the non-zero instances. Missing rain data at a monthly resolution has been recovered using a two-step hurdle analysis using only precipitation data from separate gauges by (Kim and Ahn, 2009) who use neural networks for the classification analysis and, subsequently, multivariate regression to impute daily rain. This demonstrates the suitability of ML techniques for the classification of rain events. Similarly, (Abraham and Tan, 2009), perform the two-step method on daily precipitation data and leverage covarying data using linear equations. A development of the method utilises support vector machines in the classification step but continues with linear regression in the regression step (Abraham and Tan, 2010). Teegavarapu et al. (2018) also perform a two-step method with daily precipitation, demonstrating the suitability of both weighting methods of neighbouring gauges and neural networks for data recovery. A similar two-step process is presented by Simolo et al. (2010), who estimate daily precipitation from a rain gauge network by utilising weighted probability functions to classify rain and non-rain days paired with least squares regression in the second step. A two-step analysis for the recovery of missing precipitation data has, to the best of our knowledge, not been applied to sub-hourly sampled data nor utilised state-of-the-art ML algorithms in both steps of the analysis.

In this paper, we present an approach for the recovery of missing precipitation data from a temperate oceanic climate at sub-hourly temporal resolution through a case study of a network of weather stations and a network of rain gauges in England, UK. We utilise a two-step analysis combining a binary classification step to identify the unbalanced samples of rain and no-rain, and subsequently, apply regression analysis to quantify the magnitude of rain samples when they occur. This analysis utilises and compares commonly used ML techniques capable of capturing complex nonlinear relationships including gradient boosting, nearest neighbours, bagged decision trees, neural networks and support vector machines in both steps. We investigate rain data recovery performance from analysing parameters commonly obtained from weather stations, as well as related meteorological variables, and investigate the influence of including neighbouring rain-gauge data. We consider individual analyses of each site and also the effect of pooling sites into regional analyses. Results are compared to a standard surface fitting method applied to neighbouring rain gauge data and recommendations for using ML techniques for precipitation data recovery are presented.

## 2. Methods

Here we describe the experimental setup to demonstrate the performance of our proposed framework in predicting precipitation values from different datasets.

### *2.1 Description of the data sets*

COSMOS-UK is a research-focussed environmental monitoring network developed to produce near real-time measurements of soil moisture (Evans et al., 2016). The network comprises 50 sites across the UK located to achieve a balance in representation of spatial areas, land cover and soil types, and soil moisture conditions, leading to a slightly more spatially dense network in the south and east of England. The network has been operational since 2013, with the number of sites increasing through time to 2019 (Table 2). Each site is equipped with a full range of meteorological sensors producing 30-minute resolution data. Meteorological variables are required for the calculation of soil moisture at COSMOS-UK sites, as well as for research into water and energy fluxes for which accurate and complete rainfall records are important. COSMOS-UK rainfall records were used as the target for rainfall imputation in this study. In addition to standard meteorological parameters, COSMOS-UK sites also record incoming and outgoing long- and shortwave radiation, soil moisture over a large spatial area from cosmic-ray probes, soil moisture and soil temperature from point sensors at a number (site-dependent) of locations and depths from 2-50cm , and soil heat flux at two locations at a depth of 3cm. Whilst this arrangement of sensors is specific to this network, these additional parameters are included in order to understand the potential for improving imputation accuracies from a wider range of parameters. Full details of the COSMOS-UK instrumentation and site locations can be found on the website (cosmos.ceh.ac.uk).

The Environment Agency (EA) maintains the densest network of rain gauges in England, comprising ~1200 rain gauges recording every 15 minutes, distributed across the country to provide information to operational activities and longer-term environmental understanding in areas such as water resources, flooding, and climate change. Data is available openly in near real-time over the web, whilst historic data (more than ~one year old) is available on request. Data was obtained for all EA rain gauges within a 30km radius of COSMOS-UK sites, for a period from the start of each COSMOS site to 8$^{th}$ May 2019. To match the EA data to the COSMOS-UK sampling frequency, each EA reading was summed with its preceding reading to give a total of rain for each 30-minute period. A number of EA rain gauges were used for analysis at multiple sites. Data was not obtained from equivalent networks in Scotland, Wales or Northern Ireland, which reduces the number of COSMOS-UK sites available for analysis to 37, though sites bordering England where rain gauge coverage is considered adequate are included (Table 2: Fig 1). The amount of missing rain values is 0.01 – 53.76 % of all samples within the operational time across the 37 sites. Table 2 provides summary information for COSMOS-UK sites on the length of record, proportion of missing rainfall data, proportion of time rainfall was recorded, the number of additional COSMOS-UK parameters available to be used as features for imputation, the number of available EA rain gauges within 30km, and the distance to the nearest EA rain gauge.

### *2.2 Data for ML analysis*

*Core parameters:* To represent the case of data recovery using parameters commonly recorded at weather stations, five parameters from each COSMOS-UK station are used as features to predict the precipitation, these being; air pressure, relative humidity, air temperature, wind speed, and wind direction.

*All COSMOS-UK parameters:* To test the performance of maximising parameters by including multiple meteorologically relevant variables, all parameters recorded at the COSMOS-UK sites are included as features for prediction, subject to the condition that each predictive column contains no more than 10% missing values across the operation time of the site. This results in a different number of features remaining for each site (presented in Table. 1).

*EA rain gauges:* To test the integration of rain gauges from external networks, all EA gauges within a 30km radius around each COSMOS-UK site will be added to the COSMOS-UK data as further features. As before, EA sites are not included if more than 10% of the data is missing (number of sites presented in Table. 1). These gauges will also be tested alone (no COSMOS-UK data) to observe the relative contribution of non-rain gauges data from the COSMOS-UK parameters.

*Regional model:* Similarity across sites due to geographic proximity may provide additional information for predictive modelling through pooling the data from multiple neighbouring COSMOS-UK sites to increase training examples. To test this, two regional models are tested. One region sits in the east of England, UK, and incorporates five COSMOS-UK sites (Table. 4) with a max distance of ~67 km between sites. The second region is in central England, UK and includes seven sites (Table. 4) and the max distance between sites is ~110 km. To combine the data from multiple sites, the features must be same for each site in the region. Due to different levels of missingness across the features in each site leading to different columns being removed (by the 10% missing criteria), the remaining predictors which are common to all sites in the region are retained. In addition to only the COSMOS-UK data, all the EA sites within the 30km radius surrounding each of the regional sites will be included as features. As EA gauge data was available only for the running duration of each site, the data for inclusion here is between the latest start date and earliest end date of the sites in the region allowing all neighbouring rain gauges to be included.

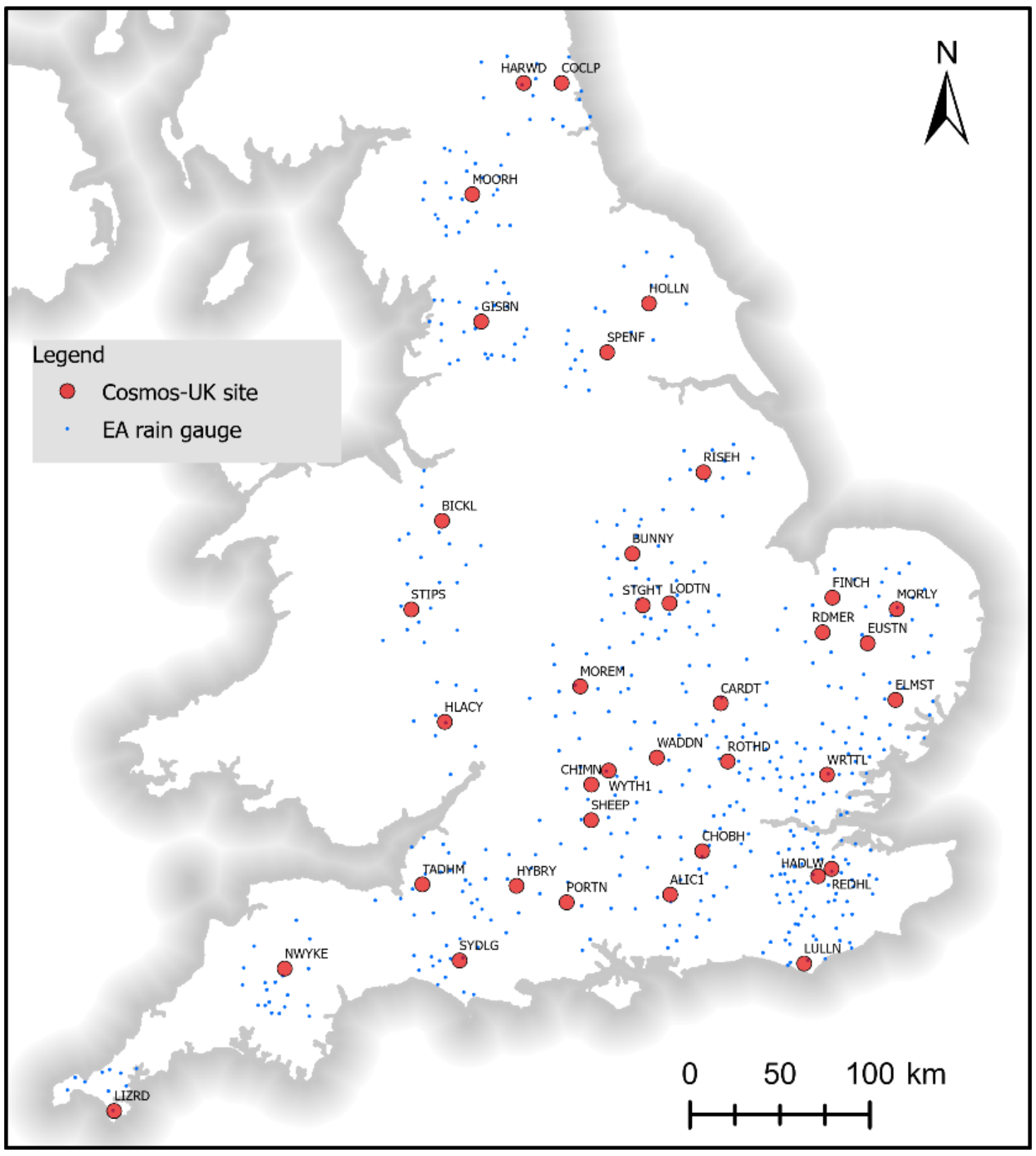

*Fig. 1. Map of England indicating COSMOS-UK monitoring sites and EA guages used in this study.*

*2.3 Machine learning techniques and grid search*

The two-step analysis (described below) finds the best performing ML model for the task of predicting missing precipitation values. Machine learning algorithms have a number of hyperparameters which are defined by the user and will optimize performance by being ‘tuned’ to the underlying data, and the task expected of the model. Indeed, using the default set of hyperparameters (itself dependant on modelling environment) is not expected to be optimal for every task. Beyond prior domain knowledge and manual experimentation, a powerful method for extracting the best hyperparameters for a given model is to iteratively search through a defined space of proposed values for each hyperparameter, judging the performance of each combination to find the optimal values (known as a grid search). Here, the data from each of the 37 COSMOS-UK sites are used to grid search a set of parameters for each ML model to optimize performance. This provides a set of values tuned to each site specifically, thus being dependant on the number of features present after removal based on missingness. The hyperparameter values being grid-searched for each analysis comes from both the default values (optimal across a range of scenarios) and the surrounding values allowing for tuning to the specific problem. For each site, a single train/test split (randomly chosen) of 70%/30% is made and subjected to each of the ML models for the task of both classification and regression. For classification, the output of the model is the precipitation column, converted to binary classes with class 0 representing a true non-rain value of 0 and class 1 being any rain value >0. The input for the models is all the remaining predictor features with performance of each set of hyperparameters based on classification accuracy. For the regression task, the output of the model is the precipitation column of every time-sample which contains a value >0 in its original unit. The input is the same time-samples across the predictor features as before. The regression task maximizes the $R^2$ score function between the predicted precipitation values and ground truth rain values. The following described models and grid searched hyperparameters are used herein:

*Gradient tree boosting:* gradient tree boosting is an ensemble method whereby multiple decision trees are iteratively fit to the data with each tree building on the previous to reduce loss and improve performance (Friedman, 2001). It is based on the principle of boosting, i.e. creating an ensemble of weak learners to deliver improved prediction accuracy. Here we use the popular implementation XGBoost (Chen and Guestrin, 2016), which uses decision tree ensembles consisting of a set of classification and regression trees (CART). XGBoost has previously shown good performance in both classification and regression with various meteorological and sensor data (Zamani Joharestani et al., 2019). Grid searched hyperparameters and values presented in Table 1.

*K-nearest neighbours:* The nearest neighbour's algorithm is a non-parametric method that can be used for both classification and regression. The output is calculated based on weighting a number of nearest neighbours in the feature space based on a distance function; most commonly Euclidean distance for continuous data. For classification, the most frequently voted label from amongst the *k* number of neighbours is assigned to the output. For regression, the average of the *k* nearest neighbours is the output. Grid searched hyperparameters and values presented in Table 1

*Random forest:* Similar to XGBoost, a random forest algorithm is also an ensemble method utilising multiple decision trees (Ho, 1995). However, rather than a sequential method for updating trees based on previous tree performance, a number of bootstrapped trees are fit in parallel with the averaged output from the trees being assigned to the output. Random forests combine Breiman's idea of bagging with the random selection of features in order to construct a collection of decision tress with controlled variance (Breiman, 2001). Grid searched hyperparameters and values presented in Table 1.

*Support vector machines:* support vector machines are algorithms which seek to find a hyperplane or a set of hyperplanes in data which is non-linearly transformed into a higher dimensional space through kernel methods (Suykens and Vandewalle, 1999). In classification, the boundary between the hyperplane and the two closest points either side of the line is maximised to distinguish between classes. For regression tasks, the hyperplane and boundary layers minimize an error function to estimate equation coefficients. Grid searched hyperparameters and values presented in Table 1.

*Neural networks:* neural networks are non-linear models and universal approximators, i.e. they can approximate any Borel measurable function from one finite dimensional space to another, provided enough hidden neurons are available (Hornik et al., 1989). Neural networks have been a very popular ML algorithm and have seen a number of improvements and milestones since the 1940s where the first attempts for creating computational models for neural networks were made. They have passed through various stages of development, such as the introduction of single layer perceptron , introduction of various activation functions, such as sigmoid, tanh, and ReLu, the use of backpropagation in neural networks , the idea of using convolution kernels with the first application on recognising hand-written digits from images published in 1989 (LeCun et al., 1989), and in the last decade the introduction of deep neural networks and deep learning, largely due to the developments and advances on graphics processing units (GPUs). Various hyperparameters need to be tuned when developing neural networks, such as the number of hidden layers, number of neurons in each layer, activation functions, dropout, and gradient descent optimization algorithms By searching through various possible values for each of these hyperparameters, one aims to optimise the tasks of classification and regression, hence improving the overall performance; this is the approach we have adopted in this study as well. For the binary stage, a single network of input layer, two hidden layers, and output layer with a sigmoid activation function is used for the binary task. In the regression step, we test three different networks of increasing depth consisting of an input layer, 2, 8, or 20 hidden layers, and an output layer with linear activation.

*Time dependence:* Capturing the time dependence of diurnal and seasonal cycles is not a solved problem in machine learning. Time units can be categorically encoded as additional features (as in) although including multiple time frames (hourly, daily, monthly) can lead to large number of additional features, increasing complexity. Furthermore, categorical encoding does not represent the similarity of neighbouring time units which are known to have a cyclical time-dependency, for example, meteorological variables such as temperature are more similar in night-hours then they are in the day. One method to preserve this cyclical nature is to represent time units as two features with one being a sine transform and the other a cosine transform (Jain et al., 2014), given by equations (1) and (2). Using a phase-offset cosine series with a sine transformation results in a 2-dimensional representation of cyclical time traits that preserves the relative similarity of proximal points in a cycle. Here, when included, control for daily and annual cycling are implemented as features, with both hours (1:24) and month (1:12) being transformed by both:

$$x_{sin} = \sin\left(\frac{2 \times \pi \times x}{\max(x)}\right) \quad (1)$$

$$x_{cos} = \cos\left(\frac{2 \times \pi \times x}{\max(x)}\right) \quad (2)$$

**Table 1.** Hyperparameters and gridsearched values for machine learning algorithms.

| Algorithm | Hyperparameters [values] |
|---|---|
| Gradient tree boosting | min_child_weight [1, 5, 10]<br>subsample [0.6, 1]<br>max_depth [8,12,16] |
| K-nearest neighbours | N_neighbours [5, 7, 9]<br>leaf_size [1, 3]<br>algorithm ['auto', 'kd_tree'] |
| Random forest | max_depth ['none', 40, 80]<br>n_estimators [100, 500, 1000]<br>min_samples_split [2, 5]<br>min_samples_leaf [1, 3] |
| Support vector machines | C [10, 100, 1000]<br>Gamma [0.0001, 0.001, 0.1, 1]<br>Kernel ['linear', 'rbf'] |

*2.4 Two-step ML imputation*

Here we present a two-step analysis for the recovery of missing rain data in which time-samples are first classified as either containing rain/not containing rain, with the absolute values for the predicted rain samples subsequently being regressed in the second step. Predictions must be compared to a known ground truth, and here all samples with missing precipitation data are discarded. As such, the remaining precipitation data from the chosen datasets is the output of the model to be predicted, with the inputs being the related sensor data from the COSMOS-UK stations and/or the EA data as described above. The two-step hurdle analysis is summarised in Fig. 2. To gauge the performance of the modelling process, k-fold cross validation is used with the average metrics of all the folds being the final measure of performance. The dataset is split into 5 folds (randomly chosen) so that each data point serves as training data four times and test data once. This ensures all data is validated in the model and minimises any bias from a single train/test split. In all cases, the test data represents unseen data for the model with the precipitation column serving as the ground truth and target of the model. Performance metrics are obtained by comparisons of the ground truth to the predictions. For each fold, the following pre-processing occurs. The input shape of data must be consistent in both training and test sets and must be complete for many ML algorithms including neural networks. To ensure completeness in the training data a complete-case sampling process is used. Each time-sample for which the data is complete (i.e. no NaNs) across all the features are retained and the rest are discarded. The training set is here not imputed for completeness as this could introduce bias in the recovered precipitation values. For the test set, discarding incomplete time-samples is not suitable because in the application of models being used to recover missing precipitation data in real time, it may be that the feature space for any one time-sample is not complete. Here, we interpolate the test set for completion using random forest models, which have previously been shown to perform well on environmental sensor data (Kim 2019). This feature imputation is an iterative process, whereby a random forest model is fitted to the data with each column in turn being the output in order of the least missing first (Stekhoven and Bühlmann, 2011). The random forest imputer is fit only to the training data to then impute the test data. This prevents information bleed and represents the case of application to new data which cannot be used to train models prior to real-time data-recovery. A common pre-processing step when using ML models is to transform the data to a consistent scale to reduce difficulty in modelling as a result of different units across the feature space. For example, at any time point, atmospheric pressure will be ~1000 hPa, where-as wind speed may be < 10 mph. This discrepancy between the scale of features can be a limitation in analyses in which Euclidean distance is calculated (as in nearest neighbour approaches) and can slow down convergence of gradient descent in deep learning. Here the data is transformed using a Min/Max method with all features being scaled between 0 and 1. Again, to prevent information bleed, the scaler is fit to just the training data and transforms both train and test set within each fold.

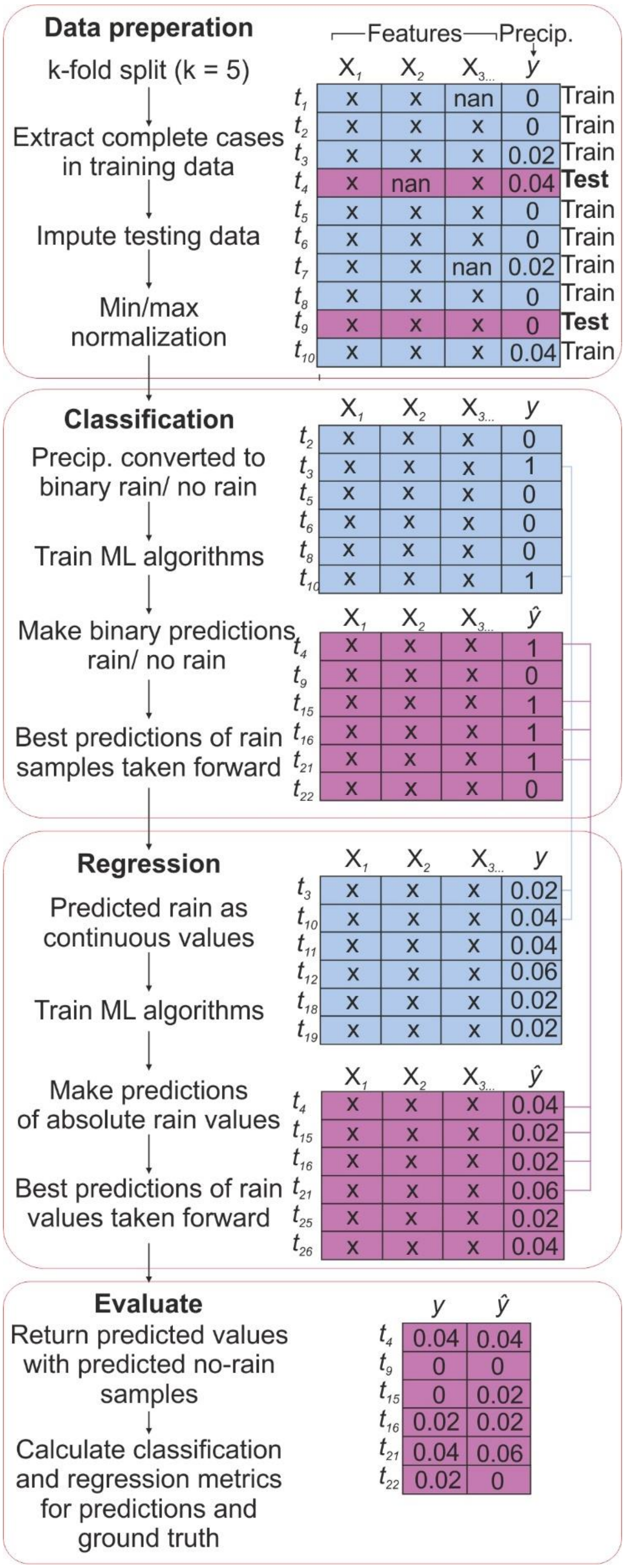

*Fig. 2. Pipeline of the two-step ML analysis. X[1…] represent the features. t[1…] represent time samples. x in tables represents a non-nan value. Blue colour: training data. Purple colour: testing data. See methods for details.*

The first step is to model each time-sample in the precipitation columns as either containing rain or being 0 (non-rain). For this, the output (precipitation) is converted to binary whereby 0 represents non-rain, and 1 represents any sample with precipitation > 0. This is done to the output of both training and test sets. The training and test data for each fold is then subject to each of the five above described ML algorithms as a binary classification task. The predictions for each fold are compared to the ground truth and classification accuracy and F1 score are calculated as:

$$F1\,score = 2\,\times \frac{Precision\,\times\,Recall}{Precision + Recall} \tag{3}$$

Where

$$Precision = \frac{TP}{TP + FP} \tag{4}$$

$$Recall = \frac{TP}{TP + FN} \tag{5}$$

Where *TP* is true positives, *FP* is false positive, *TN* is true negative, and *FN* is false negatives. The accuracy of the folds are averaged and the predictions from the model which provide the highest performance are taken through to the regression stage. F1 score is the harmonic mean of precision and recall, with the weighted F1 score being the averaged f1 score weighted by the number of samples for each class. From this step, any rain sample predicted to be class 0 (i.e. non-rain) will remain as a non-rain sample in the final predictions, and the predicted class 1 samples will be modelled for precipitation amplitude in the following regression stage.

In the regression step, the samples previously predicted to be class 1 in each fold are modelled in a regression task. From the original training data in each fold, all time-samples when the precipitation column is > 0 are retained. This discards the inflated examples of 0 from the analysis and allows the algorithms to train exclusively on examples of rain. The test set is each example in the fold which was predicted to be class 1 in the classification step. In both cases, the precipitation remains in its original units (mm). This data is then subject to each of the five ML algorithms described above with the task of recovering absolute precipitation values. As with many regression analyses, predicted values can fall outside the range of observed values. In this case, any values predicted to be < 0 are considered to be 0. The predicted values are then compared to the ground truth and the predictions from the algorithm which produces the lowest error (RMSE, see eq. 6) are retained. These predicted values are then reassigned to the sample index previously predicted as Class 1 in the classification step to produce a set of predictions for each fold including both non-rain and rain samples with absolute values. These final predictions are then compared to the ground truth with the final classification metrics and the coefficient of determination ($R^2$)and RMSE being extracted, with these being averaged across the 5 folds. The RMSE is defined as

$$\mathrm{RMSE} = \sqrt{\frac{1}{n}\sum_{i=1}^{n} e_i^2} \tag{6}$$

**Table 2.** List of sites used in this study with proportion of missing rain values and rain samples, number of COSMOS features for analysis, and number of EA gauges.

| COSMOS Site | Data duration at 30 min resolution | Missing rain values (%) | Rain samples (%) | Num. COSMOS features | Num. EA rain gauges | Distance to nearest EA site (km) |
|---|---|---|---|---|---|---|
| ALIC1 | 4y 2mo 2d 1h 30m | 0.05 | 11.16 | 27 | 19 | 3.98 |
| BICKL | 4y 3mo 9d 22h | 2.46 | 9.21 | 23 | 10 | 3.87 |
| BUNNY | 4y 3mo 11d 14h 30m | 8.46 | 8.80 | 38 | 18 | 8.33 |
| CARDT | 3y 10mo 14d 5h | 0.03 | 7.24 | 37 | 18 | 3.11 |
| CHIMN | 5y 7mo 6d 1h 30m | 16.18 | 6.74 | 35 | 13 | 8.45 |
| CHOBH | 4y 2mo 14d 1h | 1.48 | 8.48 | 35 | 20 | 3.09 |
| COCLP' | 4y 5mo 17d 0h 30m | 10.10 | 9.41 | 25 | 12 | 11.8 |
| ELMST | 2y 8mo 26d 23h 30m | 0.01 | 8.38 | 76 | 15 | 6 |
| EUSTN | 3y 1mo 7d 22h 30m | 0.04 | 9.54 | 65 | 12 | 6.87 |
| FINCH | 1y 11mo 1d 3h 30m | 0.02 | 9.52 | 67 | 9 | 11.8 |
| GISBN | 4y 8mo 23d 3h | 5.45 | 16.72 | 32 | 25 | 6.09 |
| HADLW | 2y 6mo 11d 0h 30m | 2.06 | 8.34 | 67 | 44 | 2.82 |
| HARWD | 3y 11mo 16d 3h | 3.02 | 12.87 | 26 | 10 | 1.42 |
| HLACY | 1y 27d 4h 30m | 4.49 | 7.98 | 50 | 9 | 0.74 |
| HOLLN | 5y 1mo 13d 2h 30m | 0.02 | 8.87 | 33 | 10 | 10 |
| HYBRY | 1y 8mo 22d 2h 30m | 19.19 | 15.15 | 66 | 17 | 12.6 |
| LIZRD | 4y 6mo 21d 3h | 0.02 | 13.63 | 37 | 10 | 0.72 |
| LODTN | 3y 12d | 2.14 | 11.13 | 70 | 24 | 6.07 |
| LULLN | 4y 4mo 21d 23h | 2.73 | 9.72 | 37 | 19 | 2.74 |
| MOORH | 4y 5mo 4d 3h 30m | 5.55 | 18.95 | 33 | 24 | 6.17 |
| MOREM | 5mo 22d 3h 30m | 50.05 | 14.20 | 72 | 16 | 2.94 |
| MORLY | 4y 11mo 24d 10h | 13.86 | 7.40 | 38 | 17 | 0.94 |
| NWYKE | 4y 6mo 22d 6h | 29.71 | 9.66 | 32 | 19 | 4.11 |
| PORTN | 4y 4mo 20d 4h 30m | 0.03 | 10.47 | 36 | 8 | 8.52 |
| RDMER | 3y 7mo 8d 21h 30m | 11.07 | 8.23 | 32 | 8 | 10.8 |
| REDHL | 3y 2mo 19d 23h 0m | 1.46 | 9.00 | 48 | 42 | 1.36 |
| RISEH | 3y 3d 23h | 0.56 | 8.76 | 75 | 12 | 4.73 |
| ROTHD | 4y 9mo 13d 7h | 1.54 | 8.91 | 38 | 27 | 4.93 |
| SHEEP | 5y 6mo 13d 20h 30m | 4.21 | 11.30 | 30 | 16 | 5.52 |
| SPENF | 2y 5mo 15d 5h | 0.04 | 10.06 | 67 | 12 | 14.1 |
| STGHT | 3y 8mo 19d 8h 30m | 3.52 | 8.21 | 38 | 22 | 3.9 |
| STIPS | 4y 6mo 2d' | 1.31 | 13.31 | 32 | 11 | 6.02 |
| SYDLG | 5mo 10d 3h | 53.77 | 12.98 | 69 | 16 | 2.37 |
| TADHM | 4y 6mo 24d 2h 30m | 5.82 | 9.30 | 31 | 17 | 6.01 |
| WADDN | 5y 6mo 3d 20h 30m | 7.64 | 8.81 | 31 | 12 | 6.69 |
| WRTTL | 1y 10mo 4d 4h 30m | 2.25 | 8.51 | 67 | 35 | 0.85 |
| WYTH1 | 2y 10mo 4d 8h 30m | 46.45 | 5.83 | 29 | 10 | 1.32 |

Where *n* is the number of samples with errors (*e*) between ground truth and predictions. Due to the situation whereby a sample may be predicted to be Class 1 in the classification step, but then subsequently regressed to be 0, the reported classification metrics are calculated for the final predictions for each fold. F1 score, weighted F1 score, recall, and precision is reported with classification accuracy for the evaluation of model performance in identifying the minority class of rain samples. In both classification and regression steps, the hyperparameters for each model for each site are taken from the site-specific grid search of that model. For the regional model, the data from all the included sites are represented in the randomly chosen folds. However, the final classification and regression metrics for each fold are taken on a site-by-site basis. This can then be compared to the interpolation method to observe performance of the ML model which is trained on the data from every site in the region in making predictions for each site individually.

### *2.5 Precipitation surface fitting method*

Surface fitting approaches are widely used in hydro-meteorology for the interpolation of point based observations (e.g. Li and Heap, 2011) and provide a useful benchmark to compare the ML imputation against. Here, a precipitation surface was fitted to nearby EA rain gauges using the multiquadric surface fitting technique (Hardy, 1971) in an extended form (Cole and Moore, 2008). At each COSMOS-UK site, the estimated precipitation can be expressed as a linear combination of the precipitation measurements, with weights determined by both a measure of the distance to the gauges, and also the distances between the gauges. Here, the Euclidean measure of distance is used and the "offset" parameter is set to zero to produce precipitation surfaces passing exactly through the measured value at each EA gauge (Cole and Moore, 2008). Also, the precipitation surface is assumed to be flat at a large distances, which results in unbiased estimates. Note that this form of multiquadric surface is equivalent to Kriging with a linear variogram, resulting in a close agreement between the methods for sufficiently dense rain gauge networks (Borga and Vizzaccaro, 1997).

To produce the best possible multiquadric estimates at the COSMOS-UK sites, we tested the use of different numbers of EA gauges from within 30km of each site and selected the best number of gauges to use based on the RMSE. In all cases rain gauges with weights smaller than 0.001, including negative weights, were removed and the weights re-calculated. This typically resulted in non-zero weights for 4 to 6 EA gauges (range 2 to 8). For comparison to the ML method, the predictions from the surface fitting method are converted to a binary response, for extraction of the classification metrics described above. Similarly, the $R^2$ and RMSE of the final surface-fitting predictions compared to the ground truth are also extracted. In all cases, these are taken for the same fold data from the cross-validation procedure as used in the ML process.

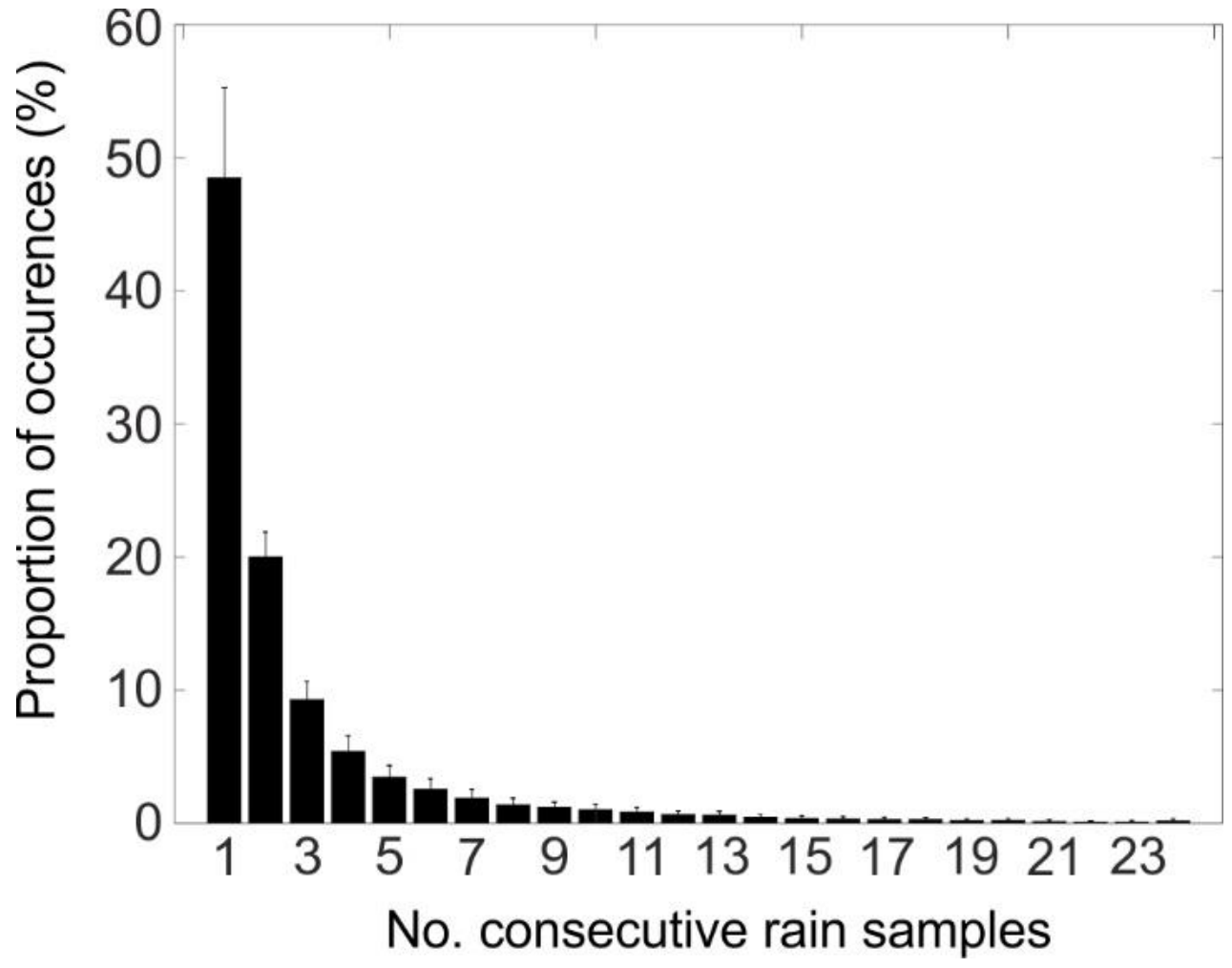

*Fig. 3. Average rain events occurring for 1 – 24 consecutive samples as a percentage of the total rain events across the 37 COSMOS sites.*

## 3. Results

The total number of COSMOS-UK features and EA gauges used are presented in Table 2. All results are presented as mean ± sd. Average number of samples which contain rain is 10.18 ± 2.78 % (range: 5.8 - 18.9 %, Table 2) representing the varying climatological conditions across the study sites. Rain presents as a single time sample in 48.48 ± 6.72 % of rain events, with consecutive samples of rain becoming increasingly uncommon (Fig. 3). Consecutive samples of rain occurring for more than 12 hours (24-time samples) are extremely rare and only occur in 0.49 ± 0.42 % of rain events across the sites. After removal based on missingness, the number of additional measured parameters available for use as features in the analysis ranges from 23 - 76 across the sites (Table 2). Correlation of the COSMOS-UK features and precipitation is weak at 30-minute sampling frequency. Of the 86 COSMOS-UK features used across the sites, correlation coefficients are between -0.155 and 0 .144. The presentation of rapidly changing precipitation from a point sensor is not reflected in the predictor variables, even in the case of meteorological variables highly associated with rain such as atmospheric pressure (correlation coefficient = -0.15) and relative humidity (correlation coefficient = 0.13; Fig. 4). Each of the above described datasets are tested using the two-step ML method. In all cases, the results from the distance-based interpolation are taken from the same k-folded data as the ML technique and can be used as a baseline performance for comparison.

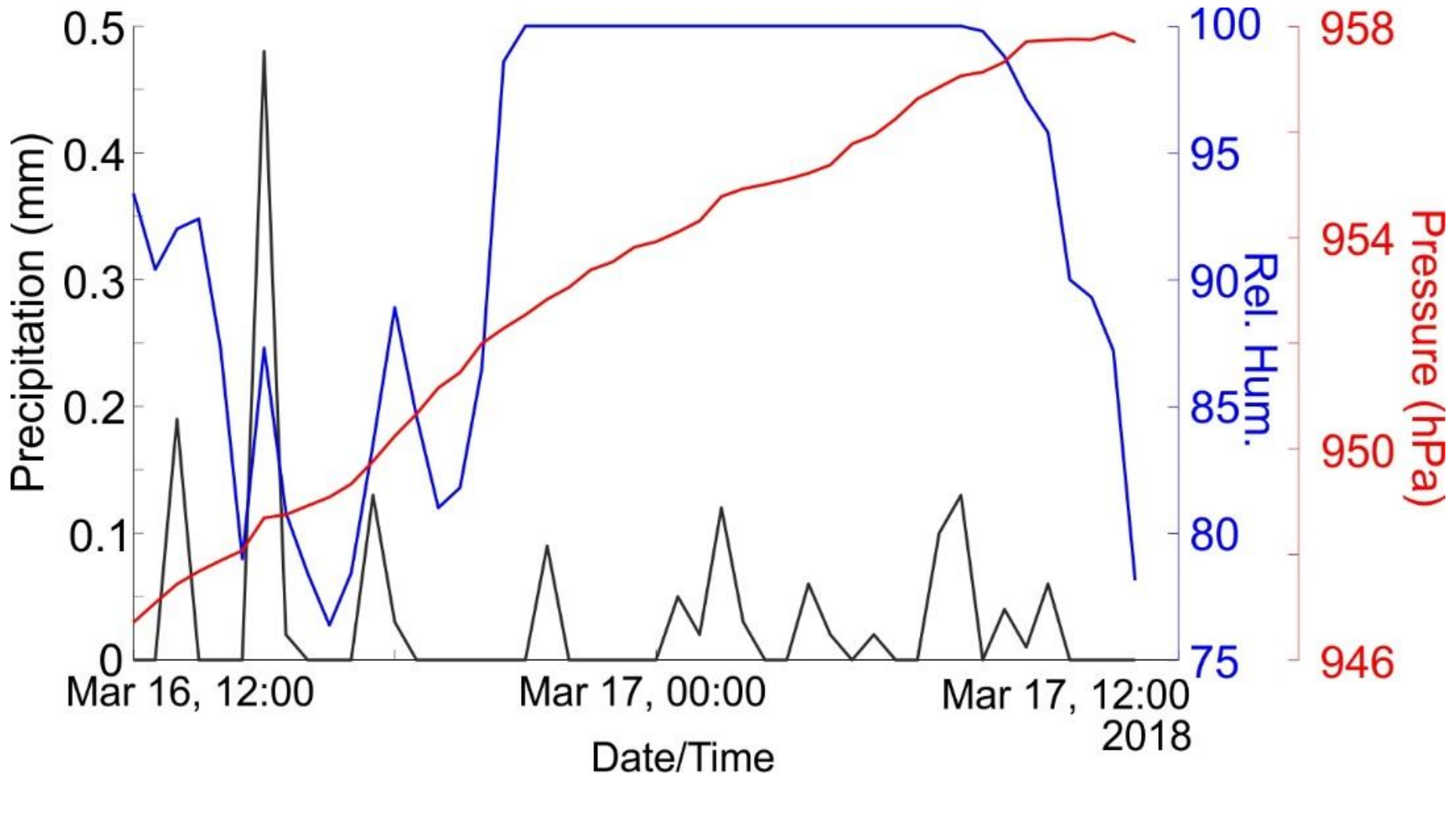

*Fig. 4. Example signal presentation of precipitation, relative humidity, and atmospheric pressure for one day from COSMOS site STIPS.*

*Surface fitting method:* Performance metrics are presented in Table 3 as the mean ± standard deviation for all sites. Boxplots for weighted f1 score, $R^2$ and RMSE are presented for each dataset analysis (Fig. 5). Here, the predictions from the interpolation method area initially converted into a binary series for comparison with the two-step method in predicting rain/non-rain samples. The interpolation process achieves an average weighted F1 score of 0.901 ± 0.044 and overall accuracy of 89.46 ± 4.07 % across the sites (Table 3). With a recall of 0.759 ± 0.112 this technique accurately predicts ~75% of rain samples, however, precision is low (0.54 ± 0.074) indicating that nearly half of all predicted rain samples are incorrect. The overestimation of the number of timesteps categorised "rain" by this method is expected because any non-zero rain value on any rain gauge with a finite weight will result in a non-zero predicted rain value. Overall prediction yielded an $R^2$ of 0.606 ± 0. and an error of 0.0151 ± 0.063 RMSE (Table 3). Predictions from the surface fitting methods are presented in Fig. 6.

*Core parameters*: This analysis concerns the six core parameters commonly found at weather station and represents a minimal number of features for data recovery. A minor improvement is noted with the inclusion of cycling control to account for daily and yearly cycles (Table 3: Fig.4). For classification, decision tree methods perform well here with the most commonly chosen model being the random forest classifier (27 sites: Fig. 7) with XGBoost the highest performing classifier for the remaining 10 sites. Average weighted F1 score across the sites was 0.909 ± 0.043 and correctly identifies just ~44% of rain (recall = 0.441 ± 0.07: Table 3). In the regression task, neural networks give the greatest performance, with the 8-hidden layer model being the best performer in 25 sites and the 2-hidden layer model in 12 sites (Fig. 7). Overall model fit is $R^2$ = 0.246 ± 068 with an error of 0.214 ± 0.06 RMSE (Table 3).

*All COSMOS-UK parameters:* This dataset represents all the data available from a comprehensive soil moisture and hydro-meteorological monitoring site, the measured parameters of which can be used for the recovery of missing precipitation data. A minor improvement in classification and regression performance is observed when controlling for cycling (Table 3: Fig.4). Decision tree methods excel here at the task of classifying rain/non-rain samples with XGBoost giving the greatest performance at

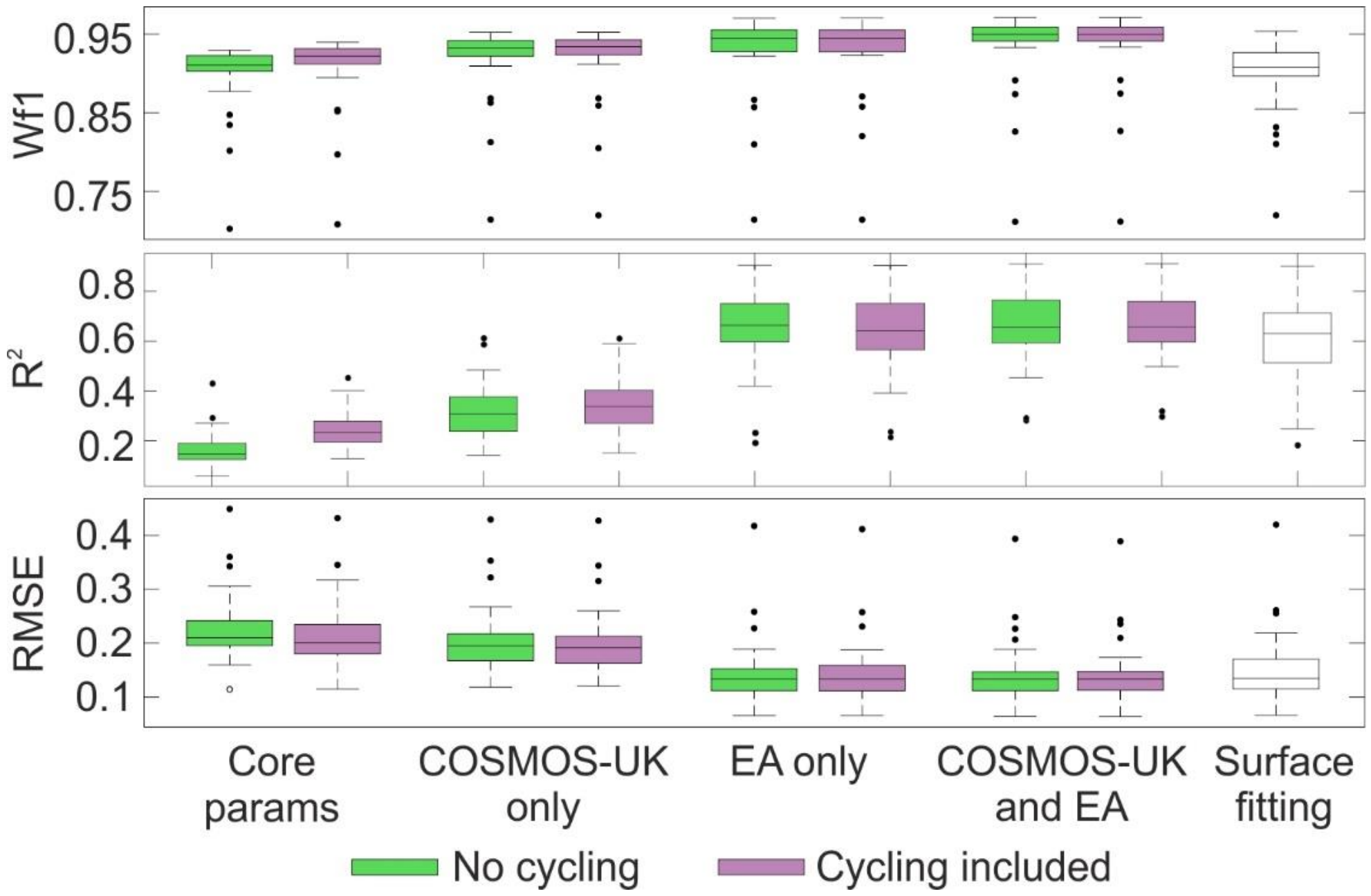

*Fig. 5. Boxplots of weighted f1, R2 and RMSE for the subsets of parameters analysed with the two-step ML method and the surface fitting process.*

**Table 3.** Averaged ± std metrics of the 37 sites for binary classification and overall regression performance for the two-step ML analysis of the datasets and the surface fitting method.

| Analysis | Cycling | Acc. | Prec. | Recall | F1 score | Weighted F1 score | $R^2$ | RMSE |
|---|---|---|---|---|---|---|---|---|
| Core params | N | 91.21 ± 3.91 | 0.733± 0.034 | 0.386± 0.073 | 0.501± 0.064 | 0.9 ± 0.06 | 0.164± 0.067 | 0.225± 0.06 |
| | Y | 91.93 ± 4 | 0.776± 0.043 | 0.441± 0.07 | 0.559± 0.061 | 0.91 ± 0.04 | 0.246± 0.068 | 0.214± 0.057 |
| COSMOS only | N | 92.68 ± 4.07 | 0.778± 0.044 | 0.537± 0.054 | 0.634± 0.048 | 0.921± 0.043 | 0.317± 0.102 | 0.203± 0.059 |
| | Y | 92.76 ± 4.06 | 0.783± 0.043 | 0.542± 0.055 | 0.639± 0.048 | 0.922± 0.043 | 0.347± 0.099 | 0.199± 0.059 |
| EA only | N | 93.69 ± 3.99 | 0.837± 0.032 | 0.597± 0.102 | 0.691± 0.085 | 0.931± 0.048 | 0.646± 0.152 | 0.144± 0.06 |
| | Y | 93.73 ± 3.92 | 0.836± 0.037 | 0.6± 0.101 | 0.694± 0.085 | 0.932± 0.047 | 0.64± 0.155 | 0.144± 0.059 |
| COSMOS and EA | N | 94.26 ± 3.88 | 0.846± 0.028 | 0.648± 0.092 | 0.73± 0.079 | 0.938± 0.047 | 0.659± 0.137 | 0.141± 0.056 |
| | Y | 94.28 ± 3.87 | 0.847± 0.027 | 0.648± 0.092 | 0.73± 0.078 | 0.938± 0.046 | 0.66± 0.131 | 0.141± 0.056 |
| Surface fitting | | 89.46 ± 4.08 | 0.54± 0.074 | 0.759± 0.112 | 0.622± 0.064 | 0.901± 0.044 | 0.606± 0.16 | 0.151± 0.06 |

29 sites with the remaining eight favouring the random forest classifier (Fig. 7). Average weighted F1 score is 0.922 ± 0.044 (Table 3). In the regression step, the most common neural networks excelled in all sites apart from two (Fig. 7). Overall model fit of the final predictions is $R^2$ = 0.347 ± 0.99 and an error 0.199 ± 0.059 RMSE (Table 3).

*EA rain gauges:* This analysis concerns the EA rain gauges surrounding each COSMOS-UK site within 30 km. The number of gauges used here is ranges from 8 to 44 across the sites (Table 2). No improvement is gained here by including cycling features (Table 3: Fig.4). For classification, ensemble decision tree methods perform the highest across 34 sites (Fig. 7). Weighted F1 score here is 0.931 ± 0.048 (Table 3). In the regression task, support vector regression was the highest performing algorithm in 26 sites, with the remaining using neural networks in 7 sites, and ensemble decision trees in four sites (Fig. 7). Model fit of the final predictions is $R^2$ = 0.646 ± 152 and an error of 0.144 ± 0.06 RMSE (Table 3).

*COSMOS-UK and EA rain gauges:* This analysis includes all the available features for each COSMOS-UK site and all the EA rain gauges surrounding the target site (number of features in Table 2). Mean distance to the closes EA rain gauge is 5.45 km (25:75 % interquartile range = 2.8:7.2 km). When including cycling features, results are minimally improved (Table 3: Fig.4). Performance here is the highest of all the datasets and represents the greatest improvement over the surface fitting interpolation method. In the classification step, ensemble decision trees perform the highest at 36 sites, with a support vector classifier favoured by the remaining site (Fig. 7). Weighted F1 score is 0.938 ± 0.046 with a classification accuracy of 94.28 ± 3.87 %. As with the surface fitting technique, this two-step analysis also underestimates the amount of rain samples (recall = 0.648 ± 0.092) yet less often misclassifies non-rain samples as containing rain (precision = 0.847 ± 0.027) (Fig. 6). For regressing these binary predictions, the highest performing algorithms are neural networks (18 sites), support vector regression (14 sites), and decision tree methods (5 sites; Fig. 6). Overall model fit is $R^2$ = 0.66 ± 0.131 with an error of 0.141 ± 0.056 RMSE (Table 3).

As the highest performing algorithm and dataset for the prediction of precipitation data we will consider the site-by-site results here in more detail, presented in Table 4. When identifying rain/non-rain samples, predictions from the two-step analysis are more accurate than those from the surface fitting method in all but one COSMOS-UK site (Table 4). For the final predictions of absolute precipitation (examples in Fig. 6), the two-step analysis offers the performing models in both overall

$R^2$ and RMSE in all but seven of the 37 sites (Table 4). For these sites in which the surface fitting method outperformed the two-step process, there is an EA rain-gauge within 7 km of the target site, suggesting the implied strong similarity between geographically proximal rain gauges (Table 4). However, this does not seem to be consistent as in the case of sites LIZRD, MORLY, and WRTTL where the two-step ML method outperforms the surface fitting method despite there being a neighbouring EA rain gauge within 1 km of the target site. One site in which surface fitting method offers better performance in classification and regression is for site MOREM for which less than 6 months of operational data was available.

*Regional model:* The two regions encompass a series of neighbouring COSMOS-UK sites and their corresponding EA rain gauges. This analysis seeks to increase the amount of training examples to the models which will then make predictions for every site individually. The CENTRAL model includes the data for seven sites (Table 5). After removal of features based on the missingness threshold of 10%, 23 common features are retained for all the sites. Additionally, 91 EA rain gauges within 30 km of any of the region sites are included. When only including the matched COSMOS-UK features common to the region sites, (no EA gauges) the predictive performance is not as high as the single site two-step method (Table 4). The random forest classifier performs the highest here in classification with neural networks providing the lowest error in final predictions.

**Table 4.** Metrics for classification and regression analysis for each individual site from the best performing method (see text for details). All results are average ± std for the five folds of the cross validation. Results for the surface fitting method for the same site and folds are included. Bracketed number next to NN indicate number of hidden layers. Class. = classification analysis. Reg. = regression analysis.

| Site | Two-step ML | | | | | Surface fitting | | |
|---|---|---|---|---|---|---|---|---|
| | Class. | Weighted F1 score | Reg. | $R^2$ | RMSE | Weighted F1 score | $R^2$ | RMSE |
| ALIC1 | XGB | 0.959 ± 0.002 | SVR | 0.785 ± 0.054 | 0.113 ± 0.027 | 0.926 ± 0.005 | 0.777 ± 0.043 | 0.115 ± 0.024 |
| BICKL | XGB | 0.95 ± 0.001 | NN(2) | 0.593 ± 0.035 | 0.133 ± 0.01 | 0.9 ± 0.008 | 0.737 ± 0.026 | 0.107 ± 0.007 |
| BUNNY | XGB | 0.949 ± 0.002 | NN(2) | 0.658 ± 0.052 | 0.113 ± 0.012 | 0.921 ± 0.011 | 0.633 ± 0.038 | 0.117 ± 0.011 |
| CARDT | RF | 0.963 ± 0.001 | SVR | 0.726 ± 0.044 | 0.088 ± 0.009 | 0.932 ± 0.014 | 0.65 ± 0.065 | 0.099 ± 0.004 |
| CHIMN | XGB | 0.966 ± 0.001 | NN(2) | 0.52 ± 0.115 | 0.158 ± 0.048 | 0.92 ± 0.01 | 0.458 ± 0.107 | 0.168 ± 0.046 |
| CHOBH | XGB | 0.971 ± 0.001 | SVR | 0.815 ± 0.034 | 0.098 ± 0.015 | 0.941 ± 0.008 | 0.637 ± 0.32 | 0.124 ± 0.038 |
| COCLP | XGB | 0.945 ± 0.001 | NN(2) | 0.583 ± 0.049 | 0.126 ± 0.016 | 0.878 ± 0.014 | 0.544 ± 0.043 | 0.132 ± 0.014 |
| ELMST | XGB | 0.957 ± 0.001 | XGB | 0.668 ± 0.044 | 0.112 ± 0.013 | 0.924 ± 0.014 | 0.706 ± 0.012 | 0.106 ± 0.012 |
| EUSTN | XGB | 0.941 ± 0.001 | NN(2) | 0.539 ± 0.129 | 0.174 ± 0.053 | 0.905 ± 0.009 | 0.478 ± 0.124 | 0.185 ± 0.052 |
| FINCH | XGB | 0.952 ± 0.001 | NN(2) | 0.553 ± 0.047 | 0.142 ± 0.018 | 0.918 ± 0.008 | 0.528 ± 0.068 | 0.146 ± 0.02 |
| GISBN | XGB | 0.941 ± 0.003 | SVR | 0.789 ± 0.028 | 0.164 ± 0.017 | 0.876 ± 0.003 | 0.669 ± 0.068 | 0.205 ± 0.017 |
| HADLW | RF | 0.967 ± 0.002 | SVR | 0.807 ± 0.059 | 0.113 ± 0.03 | 0.943 ± 0.014 | 0.821 ± 0.058 | 0.109 ± 0.03 |
| HARWD | XGB | 0.938 ± 0.001 | NN(2) | 0.69 ± 0.085 | 0.21 ± 0.037 | 0.898 ± 0.006 | 0.66 ± 0.072 | 0.219 ± 0.024 |
| HLACY | XGB | 0.964 ± 0.002 | SVR | 0.911 ± 0.017 | 0.064 ± 0.009 | 0.946 ± 0.008 | 0.902 ± 0.027 | 0.067 ± 0.011 |
| HOLLN | XGB | 0.958 ± 0.001 | SVR | 0.499 ± 0.112 | 0.173 ± 0.036 | 0.908 ± 0.007 | 0.448 ± 0.135 | 0.181 ± 0.038 |
| HYBRY | XGB | 0.874 ± 0.005 | NN(2) | 0.658 ± 0.047 | 0.118 ± 0.02 | 0.822 ± 0.016 | 0.589 ± 0.063 | 0.129 ± 0.02 |
| LIZRD | XGB | 0.941 ± 0.001 | SVR | 0.764 ± 0.062 | 0.137 ± 0.029 | 0.915 ± 0.007 | 0.654 ± 0.082 | 0.166 ± 0.036 |
| LODTN | XGB | 0.941 ± 0.002 | NN(2) | 0.657 ± 0.069 | 0.109 ± 0.014 | 0.905 ± 0.008 | 0.612 ± 0.063 | 0.116 ± 0.012 |
| LULLN | XGB | 0.959 ± 0.001 | SVR | 0.782 ± 0.073 | 0.128 ± 0.028 | 0.945 ± 0.013 | 0.76 ± 0.065 | 0.134 ± 0.024 |
| MOORH | XGB | 0.891 ± 0.003 | NN(2) | 0.297 ± 0.058 | 0.389 ± 0.051 | 0.832 ± 0.004 | 0.183 ± 0.034 | 0.42 ± 0.053 |
| MOREM | SVC | 0.712 ± 0.026 | SVR | 0.712 ± 0.056 | 0.074 ± 0.022 | 0.72 ± 0.027 | 0.749 ± 0.05 | 0.069 ± 0.018 |
| MORLY | XGB | 0.969 ± 0.001 | XGB | 0.738 ± 0.131 | 0.134 ± 0.046 | 0.953 ± 0.003 | 0.491 ± 0.283 | 0.178 ± 0.029 |
| NWYKE | XGB | 0.933 ± 0.002 | NN(2) | 0.655 ± 0.042 | 0.136 ± 0.015 | 0.855 ± 0.004 | 0.522 ± 0.078 | 0.159 ± 0.015 |
| PORTN | XGB | 0.946 ± 0.001 | SVR | 0.644 ± 0.06 | 0.134 ± 0.019 | 0.908 ± 0.004 | 0.6 ± 0.084 | 0.142 ± 0.023 |
| RDMER | XGB | 0.949 ± 0.002 | NN(8) | 0.319 ± 0.112 | 0.243 ± 0.041 | 0.917 ± 0.009 | 0.249 ± 0.061 | 0.255 ± 0.031 |
| REDHL | RF | 0.961 ± 0.002 | RF | 0.843 ± 0.066 | 0.088 ± 0.025 | 0.928 ± 0.006 | 0.88 ± 0.042 | 0.076 ± 0.01 |
| RISEH | XGB | 0.96 ± 0.002 | RF | 0.617 ± 0.092 | 0.132 ± 0.036 | 0.925 ± 0.007 | 0.566 ± 0.074 | 0.141 ± 0.033 |
| ROTHD | XGB | 0.958 ± 0.001 | NN(2) | 0.64 ± 0.191 | 0.145 ± 0.081 | 0.929 ± 0.007 | 0.653 ± 0.178 | 0.143 ± 0.079 |
| SHEEP | XGB | 0.944 ± 0.002 | NN(2) | 0.639 ± 0.065 | 0.134 ± 0.01 | 0.907 ± 0.008 | 0.631 ± 0.085 | 0.135 ± 0.016 |
| SPENF | XGB | 0.946 ± 0.001 | NN(2) | 0.529 ± 0.139 | 0.146 ± 0.044 | 0.905 ± 0.007 | 0.453 ± 0.08 | 0.159 ± 0.041 |
| STGHT | XGB | 0.956 ± 0.005 | NN(2) | 0.758 ± 0.097 | 0.109 ± 0.019 | 0.89 ± 0.035 | 0.746 ± 0.096 | 0.112 ± 0.016 |
| STIPS | XGB | 0.937 ± 0.002 | NN(2) | 0.613 ± 0.047 | 0.153 ± 0.011 | 0.894 ± 0.006 | 0.299 ± 0.601 | 0.189 ± 0.058 |
| SYDLG | XGB | 0.827 ± 0.005 | XGB | 0.88 ± 0.081 | 0.113 ± 0.023 | 0.81 ± 0.018 | 0.851 ± 0.047 | 0.133 ± 0.01 |
| TADHM | XGB | 0.951 ± 0.001 | SVR | 0.7 ± 0.017 | 0.124 ± 0.004 | 0.906 ± 0.007 | 0.674 ± 0.039 | 0.129 ± 0.01 |
| WADDN | XGB | 0.949 ± 0.002 | NN(2) | 0.599 ± 0.042 | 0.121 ± 0.013 | 0.897 ± 0.005 | 0.604 ± 0.031 | 0.121 ± 0.014 |
| WRTTL | XGB | 0.958 ± 0.002 | NN(2) | 0.642 ± 0.228 | 0.14 ± 0.072 | 0.934 ± 0.011 | 0.466 ± 0.393 | 0.157 ± 0.068 |
| WYTH1 | XGB | 0.939 ± 0.004 | NN(2) | 0.616 ± 0.099 | 0.236 ± 0.045 | 0.911 ± 0.017 | 0.542 ± 0.05 | 0.261 ± 0.049 |

Performance of the regional model is improved by including all the EA rain gauges, however, predictions are still more accurate from the single site model (Table 5). When including the EA gauge data, XGBoost performs the highest in classification with neural networks still favoured in regression. For the EAST region, 31 common COSMOS-UK features are retained with 40 EA sites. As before, performance is improved when including the EA sites, but predictions are not as accurate as those of the single site analysis (Table 5).

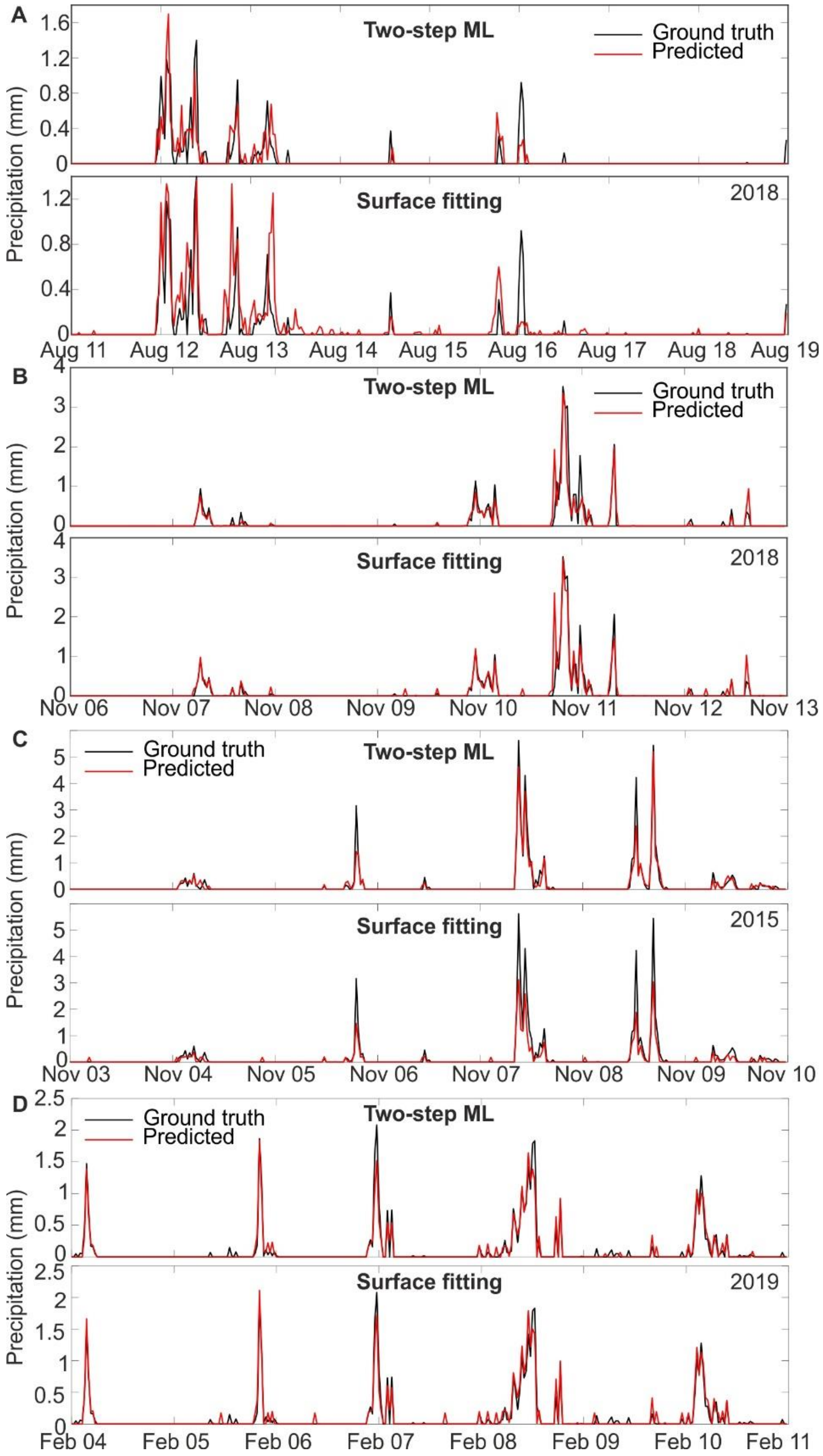

*Fig. 6. Example predictions and ground truth from both two-step ML method and the surface fitting method for sites A. COCLP and B. WRTTL. C. HARWD and D. HLACY. Note reduction of incorrect small amplitude rain predictions in the two-step ML method.*

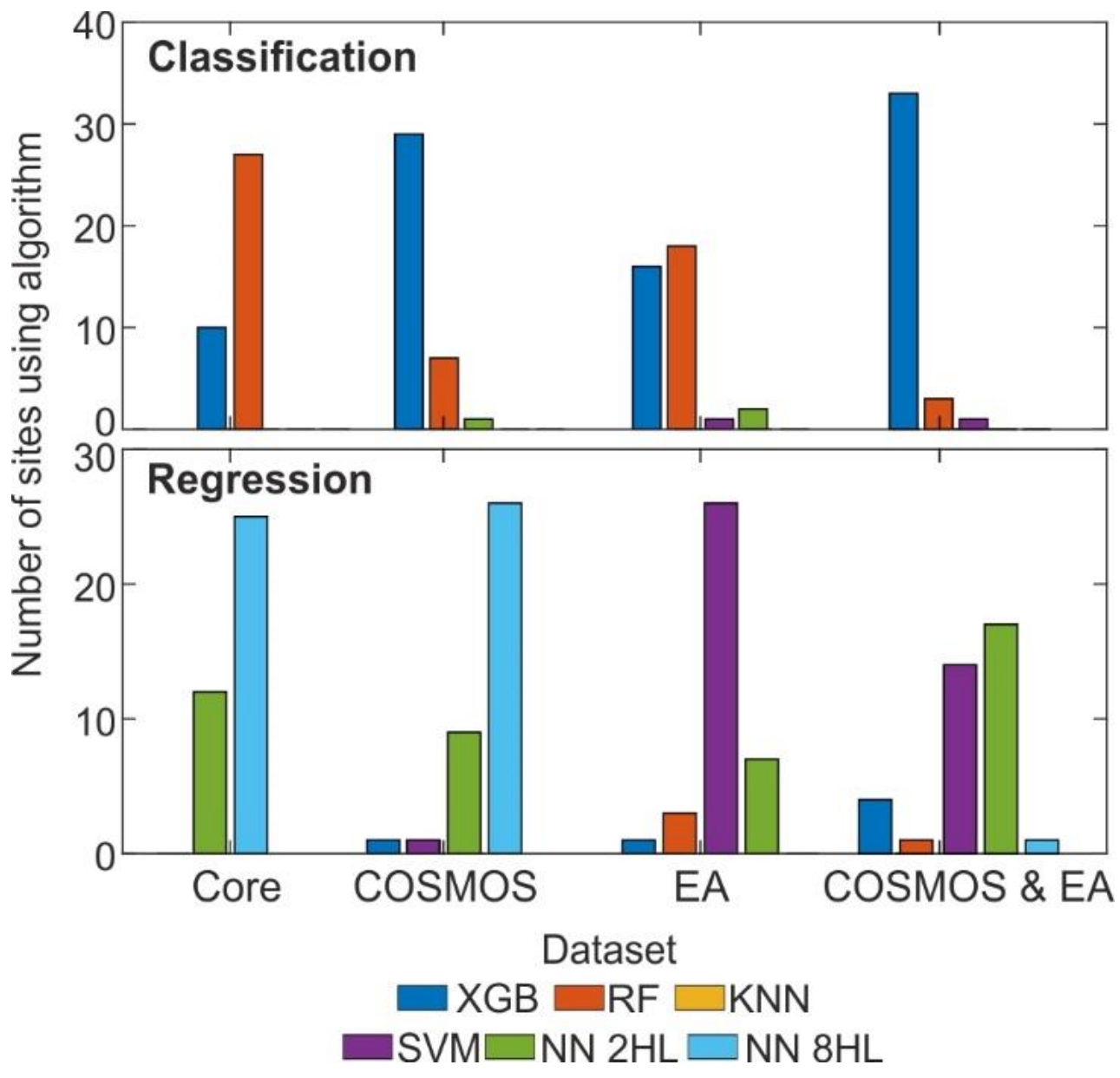

*Fig. 7. Prevalence of the best performing ML algorithms across the datasets analysed in this study. Note ensemble decision trees are favoured for classification while neural networks and support vector machines are dominate for regression.*

## 4. Discussion

In this study we present a two-step process utilising current ML techniques which here outperforms a surface fitting technique for the recovery of missing precipitation data at 30-minute resolution. This technique is easily implemented and customizable, and searches through numerous ML algorithms to find the best performance for a given problem. A primary benefit of the two-step ML process is to reduce the number of small-scale incorrect predictions of the surface fitting method (i.e. the misclassification of a non-rain sample). This may be due to the prevalence of local convective rain events which will not encompass even closely spaced rain gauges leading to erroneous results when weighting the output of neighbouring gauges for a target site. Data recovery is possible when only a limited number of parameters are available (such as the core parameters analysis), particularly the discrimination of rain or non/rain samples. When including further parameters from the COSMOS-UK sites performance is expectedly improved. Among these extra features is soil moisture (volumetric water content) suggesting that the addition of soil moisture sensors, which are cheap and easy to install in monitoring sites, may support future requirements for precipitation data recovery. The two-step ML analysis of only the EA gauges surrounding the target site perform better than the surface fitting method for the recovery of data, with inclusion of the available meteorological variables increasing performance further. This demonstrates that ML models can extract additional information from noisy features and can process the large scale and dimensionality of meteorological variables recorded at high sampling frequencies. Interestingly, performance is also increased in certain cases when a neighbouring rain gauge is within 1 km of the target site. The increased predictive performance of ML algorithms with increasing data agrees with previous works (Chen et al., 2019) and this represents a promising line of enquiry for data recovery. The use of ML models such as those presented here can also be used in quality control measures applied after data collection.

In general terms, ensemble decision tree methods (XGBoost and random forest) perform well in the classification task of rain/non-rain, while neural networks perform well in the regression stage. However, some variation is observed demonstrating that not all ML algorithms are equal on datasets which are superficially similar but may vary widely in predictive power due to differing number and types of features available for analysis. This also highlights the variation in ML algorithm mechanisms, despite them all being capable of extracting information from non-linear and noisy data sets.

**Table 5.** Classification and regression results for the predictions of each individual site from the regional analysis. All results are average ± std for the five folds of the cross validation.

| Region | EA gauges included | Site | Weighted F1 score | $R^2$ | RMSE |
|---|---|---|---|---|---|
| Central | No | ALIC1 | 0.938 ± 0.002 | 0.393 ± 0.058 | 0.194 ± 0.02 |
| | | CARDT | 0.945 ± 0.001 | 0.263 ± 0.042 | 0.145 ± 0.01 |
| | | CHIMN | 0.945 ± 0.001 | 0.188 ± 0.09 | 0.214 ± 0.04 |
| | | CHOBH | 0.944 ± 0.002 | 0.373 ± 0.077 | 0.183 ± 0.014 |
| | | ROTHD | 0.937 ± 0.002 | 0.279 ± 0.074 | 0.204 ± 0.065 |
| | | SHEEP | 0.916 ± 0.002 | 0.226 ± 0.036 | 0.186 ± 0.006 |
| | | WADDN | 0.923 ± 0.002 | 0.221 ± 0.079 | 0.162 ± 0.014 |
| Central | Yes | ALIC1 | 0.951 ± 0.001 | 0.607 ± 0.046 | 0.156 ± 0.021 |
| | | CARDT | 0.96 ± 0.002 | 0.613 ± 0.065 | 0.105 ± 0.013 |
| | | CHIMN | 0.963 ± 0.001 | 0.388 ± 0.16 | 0.186 ± 0.048 |
| | | CHOBH | 0.968 ± 0.001 | 0.622 ± 0.073 | 0.143 ± 0.026 |
| | | ROTHD | 0.954 ± 0.002 | 0.41 ± 0.175 | 0.185 ± 0.074 |
| | | SHEEP | 0.937 ± 0.002 | 0.542 ± 0.061 | 0.143 ± 0.01 |
| | | WADDN | 0.936 ± 0.002 | 0.432 ± 0.073 | 0.138 ± 0.016 |
| East | No | ELMST | 0.942 ± 0.003 | 0.24 ± 0.069 | 0.192 ± 0.063 |
| | | EUSTN | 0.91 ± 0.009 | 0.287 ± 0.126 | 0.244 ± 0.075 |
| | | FINCH | 0.936 ± 0.002 | 0.371 ± 0.093 | 0.186 ± 0.031 |
| | | MORLY | 0.942 ± 0.002 | 0.329 ± 0.11 | 0.21 ± 0.057 |
| | | RDMER | 0.926 ± 0.006 | 0.238 ± 0.071 | 0.287 ± 0.078 |
| East | Yes | ELMST | 0.948 ± 0.003 | 0.436 ± 0.032 | 0.163 ± 0.042 |
| | | EUSTN | 0.923 ± 0.007 | 0.414 ± 0.168 | 0.223 ± 0.087 |
| | | FINCH | 0.943 ± 0.003 | 0.382 ± 0.124 | 0.182 ± 0.019 |
| | | MORLY | 0.959 ± 0.004 | 0.636 ± 0.041 | 0.155 ± 0.041 |
| | | RDMER | 0.935 ± 0.002 | 0.276 ± 0.117 | 0.273 ± 0.048 |

Interestingly, in no cases was the k-nearest neighbour approach utilised for either classification or regression, likely due to the complex and weakly correlated relationship between predictor features and target. Similarly, the deep neural network of 20 hidden layers was not utilised, possibly due to overfitting with the number of training examples used here. When using a small number of parameters which may not capture time-dependency, performance is improved by explicitly controlling for the daily and annual cycling of time through the cos/sin technique. This effect is reduced when using multiple different meteorological parameters where it is either that (1) the effect of daily and annual time dependence is represented in the parameters themselves, or (2) the explanatory power of cycling control at sub-hourly scale is not sufficient to improve the performance in the presence of more highly dimensional data.

At 30-minute resolution, the prevalence of rain events occurring as single time samples represents a challenge for the accurate recovery of precipitation data as correlation with predictor variables is low. In simple terms, should a rain event occur for a single 30-minute sample, the subsequent sample will not contain precipitation while the predictive features (such as relative humidity or soil moisture) will still reflect the conditions of 'rain' from the previous sample due to slow rates of change therein (Fig. 4). This noise in the predictor variables makes the classification of unbalanced non-rain and rain samples challenging, with both the two-step ML analysis and the surface fitting technique underestimating the number of samples which contain rain. Of note is that the meteorological variables in this analysis come from point sensors positioned at the target site, or neighbouring sites for EA rain gauges. Inclusion of data from other sources such as gridded cloud cover from satellite observations, or rainfall radar data ( Moore et al., 2004; Theeuwes et al., 2019) could improve predictive performance of precipitation data. Increasing predictor features may require a process of feature selection (as in Zamani Joharestani et al., 2019) to improve performance if the space becomes highly dimensional. Beyond including further features, increasing training examples from combining multiple sites may not improve performance as our regional model demonstrates. This agrees with previous hydrological modelling on grouped regions of single sites (Kratzert et al., 2018) and is likely due to the geographical spread of the sites and the varying small-scale climatic conditions surrounding each location.

A further line of enquiry for the recovery of time-series precipitation data is sequence analysis. In hydrological modelling, the use of wavelet transformations to capture intrinsic time-dependency with ML methods has produced promising results (reviewed in (Nourani et al., 2014). Similarly, Long Short-Term Memory (LSTM) recurrent neural networks (Hochreiter and Schmidhuber, 1997) are capable of capturing long term dependencies in hydrological time series data (Kratzert et al., 2018). However, as highlighted in Nourani et al. (2014), the auto regressive properties of precipitation are more significant at larger time scales (monthly and above) with correlation with predictor variables weak at higher resolutions as shown here. Accordingly, time-series analyses at daily resolution are fewer in the literature and are focussed on the task of forecasting (Kisi and Cimen, 2012). The predictive capabilities of sequence analyses at hourly and sub-hourly time scales for data recovery, particularly including additional hydro-meteorological parameters, deserves further attention.

## 5. Conclusions

Here, we demonstrate the performance benefit of ML techniques for recovering precipitation data. A key benefit of the proposed two-step ML method is its potential for deployment for real-time gap filling. Models can be trained offline and be routinely updated as new data becomes available within a user-defined schedule (e.g. monthly or bi-annually). These models can then be deployed with the associated pre-processing steps for near instantaneous recovery of missing values as they occur. The use of a defined target-neighbour distance for data inclusion with publicly available API-addressable data sources (such as the Environmental Agency, UK) can help automation of data extraction and analysis. Single site models, where possible, can improve over generalisable single architectures (as in (Cramer et al., 2017) although the number of target sites and computational cost must be taken into account when developing pipelines for data recovery.

## Author contributions

GL and MF conceived the project, with all authors contributing to evolution of the research goals. GL and BC developed and implemented the machine learning methodology. OS, SC, JE, SS, and JW implemented the surface fitting methodology. MF and OS provided the COSMOS-UK and EA data. All authors contributed to the final version of the manuscript.

## Competing interests

The authors declare no competing interests.

## Acknowledgements

We thank the Environmental Agency for providing requested data.

## Funding

This research was supported by a UKRI-NERC Strategic Priority grant “Engineering Transformation for the Integration of Sensor Networks: A Feasibility Study” (NE/S016236/1 & NE/S016244/1).